\documentclass[sigconf]{aamas}

\usepackage{balance} 
\usepackage{array}
\usepackage{subcaption}
\usepackage{soul}
\usepackage{hyperref}

\doi{VZXJ8026}

\makeatletter
\gdef\@copyrightpermission{
  \begin{minipage}{0.2\columnwidth}
   \href{https://creativecommons.org/licenses/by/4.0/}{\includegraphics[width=0.90\textwidth]{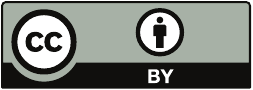}}
  \end{minipage}\hfill
  \begin{minipage}{0.8\columnwidth}
   \href{https://creativecommons.org/licenses/by/4.0/}{This work is licensed under a Creative Commons Attribution International 4.0 License.}
  \end{minipage}
  \vspace{5pt}
}
\makeatother

\setcopyright{ifaamas}
\acmConference[AAMAS '26]{Proc.\@ of the 25th International Conference
on Autonomous Agents and Multiagent Systems (AAMAS 2026)}{May 25 -- 29, 2026}
{Paphos, Cyprus}{C.~Amato, L.~Dennis, V.~Mascardi, J.~Thangarajah (eds.)}
\copyrightyear{2026}
\acmYear{2026}
\acmDOI{}
\acmPrice{}
\acmISBN{}

\title{Enhancing Goal Inference via Correction Timing}

\author{Anjiabei Wang}
\orcid{0009-0008-4345-333X}
\affiliation{
  \institution{Yale University}
  \city{New Haven, CT}
  \country{United States}}
\email{anjiabei.wang@yale.edu}

\author{Shuangge Wang}
\orcid{0009-0004-1293-5302}
\affiliation{
  \institution{Yale University}
  \city{New Haven, CT}
  \country{United States}}
\email{shuangge.wang@yale.edu}

\author{Tesca Fitzgerald}
\orcid{0000-0003-0867-0546}
\affiliation{
  \institution{Yale University}
  \city{New Haven, CT}
  \country{United States}}
\email{tesca.fitzgerald@yale.edu}

\begin{abstract}
Corrections offer a natural modality for people to provide feedback to a robot, by (i) intervening in the robot’s behavior when they believe the robot is failing (or will fail) the task objectives and (ii) modifying the robot’s behavior to successfully fulfill the task. Each correction offers information on what the robot should and should not do, where the corrected behavior is more aligned with task objectives than the original behavior. Most prior work on learning from corrections involves interpreting a correction as a new demonstration (consisting of the modified robot behavior), or a preference (for the modified trajectory compared to the robot’s original behavior). However, this overlooks one essential element of the correction feedback, which is the human’s decision to intervene in the robot’s behavior in the first place. This decision can be influenced by multiple factors including the robot’s task progress, alignment with human expectations, dynamics, motion legibility, and optimality. In this work, we investigate whether the timing of this decision can offer a useful signal for inferring these task-relevant influences. In particular, we investigate three potential applications for this learning signal: (1) identifying features of a robot’s motion that may prompt people to correct it, (2) quickly inferring the final goal of a human’s correction based on the timing and initial direction of their correction motion, and (3) learning more precise constraints for task objectives. Our results indicate that correction timing results in improved learning for the first two of these applications. Overall, our work provides new insights on the value of correction timing as a signal for robot learning.
\end{abstract}


\newcommand{\BibTeX}{\rm B\kern-.05em{\sc i\kern-.025em b}\kern-.08em\TeX}

\begin{document}


\pagestyle{fancy}
\fancyhead{}


\maketitle 


\section{Introduction}

When a robot is performing a task and appears to be failing (or about to fail), a human can intervene and physically correct the robot's motion to achieve the intended task objective. Compared to traditional paradigms with designated teaching and deployment cycles, these intervention-based interactions offer a practical source of training data: 1) they are more data-efficient \citep{hoque2023interventional}, as the robot is continuously deployed and corrected only when necessary; and 2) they enable more intuitive interactions \citep{bajcsy2018learning, losey2022physical}, since they draw on human domain expertise without requiring technical interfaces. In practice, people provide corrections due to perceived violations of task constraints, such as task progress \citep{atkeson1997robot,kumra2021learning}, safety constraints \citep{ak2023learning, saunders2017trial}, or user preferences \citep{wilde2018learning, wilde2020improving}, making them potentially highly informative toward learning these task constraints. 

Prior work on learning from corrections \cite{kalakrishnan2013learning, bajcsy2018learning, ratliff2006maximum, jain2015learning, karlsson2017autonomous} primarily formulates it as an Inverse Reinforcement Learning (IRL) problem, where the robot infers the reward function it should optimize during the task by treating corrections as evidence about the reward function's parameters.
Within the IRL framework, prior work has focused on the spatial aspects of corrections — such as where and how the robot’s motion is adjusted — while overlooking another critical dimension: the human’s decision to intervene in the robot's behavior in the first place. Prior studies show that motion features such as efficiency, safety, legibility, and human-likeness shape how people interpret and respond to robots \citep{bajcsy2017learning, jain2020anticipatory, maurice2017velocity, ak2023learning, saunders2017trial, dragan2013legibility}. 
Based on the interpretations on robot motions, humans may decide that the robot’s current trajectory is inadequate and intervene to ensure task success. Thus, the decision to intervene reflects the moment when humans internally judge that the robot requires assistance — and provides insight into \emph{why} the correction occurs \citep{tian2023matters, korkmaz2025mile}.

In this work, we focus on physical correction to a robot. We \textbf{hypothesize that the timing of human's intervention decision offers insight into the underlying task objectives}. We investigate this hypothesis in the context of three learning-related applications (and corresponding research questions):

\textbf{RQ1}: What features of the robot's motion prompt people to correct it? 
We aim to understand when and why corrections occur, enabling the future design of robot trajectories that either elicit informative corrections or avoid unnecessary ones. 

\textbf{RQ2}: Can we utilize the information available at the onset of the correction to directly infer the task goal?
This capability could enable future robots to respond to human corrections in real time.

\textbf{RQ3}: Can we learn more precise constraints about the task goal by using timing information? 
This would allow a robot to learn more precise task constraint information that may not be captured from spatial cues alone. 

Our results indicate that timing information contributes meaningfully to the first two applications. Our main contributions include:
\textbf{1)} We evaluate the contribution of individual trajectory-based features to correction timing prediction through a feature ablation study, providing insights into which aspects of robot motion influence humans’ decisions to intervene.
\textbf{2)} We demonstrate that incorporating timing information enhances early inference of human-intended goals from correction onset cues, but provides limited gains for refining fine-grained task constraint learning.

\begin{figure}
    \centering
    \includegraphics[width=1.0\linewidth, trim=0 0 450 0, clip]{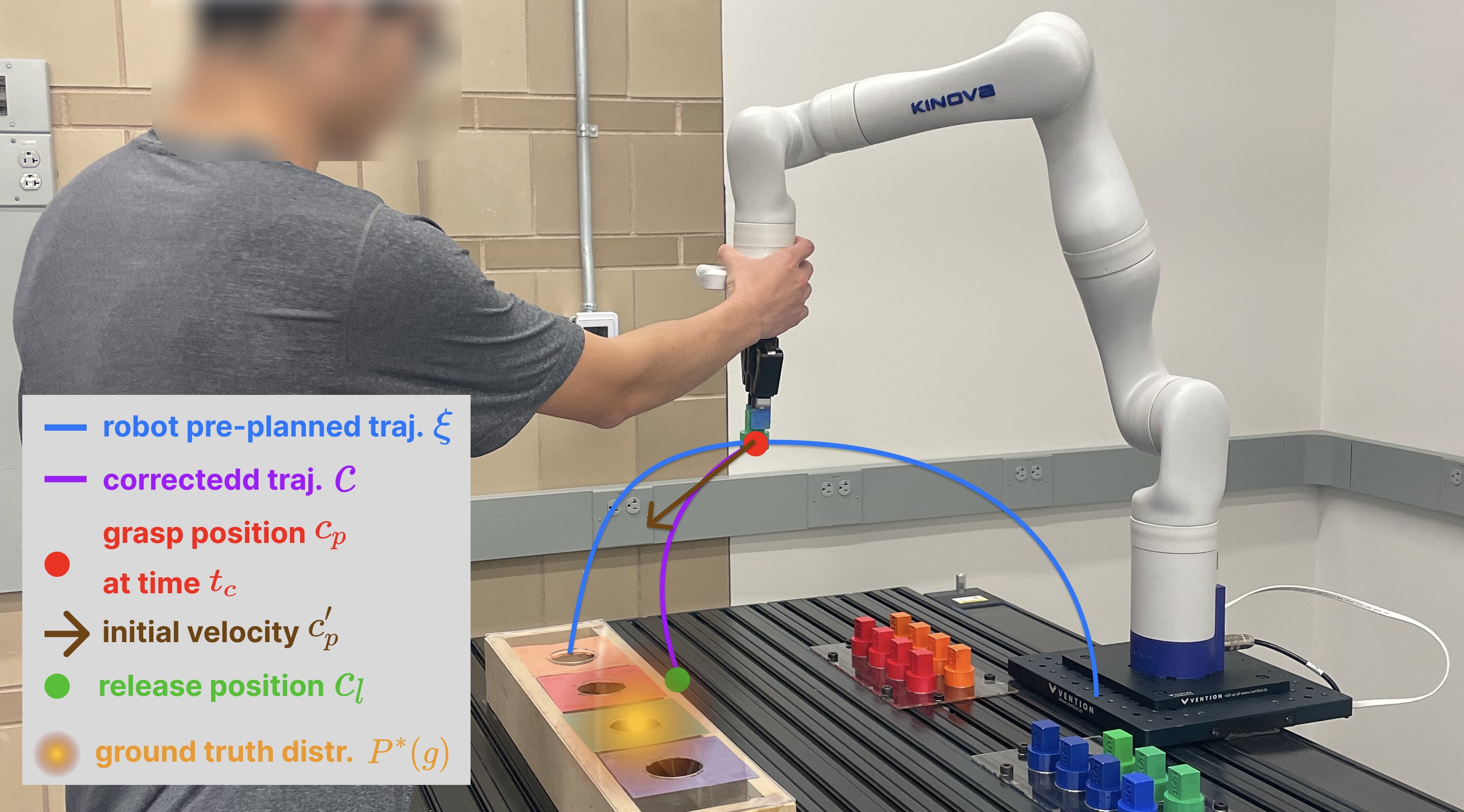}
    \caption{A human intervenes during a robot’s pre-planned trajectory $\xi$, resulting in a corrected trajectory $c$. The timing information is represented by $t_c$, while the spatial information includes  $c_p$, $c_p'$, and $c_l$. }
    \label{fig:correction_def}
\end{figure}

\section{Related Work}

\subsection{Learning from Corrections}

Correction feedback occurs when a human intervenes during a robot’s task execution to modify its ongoing motion and help it succeed. Prior work has leveraged such corrections as demonstrations. For instance, \citet{zhang2019learning} treat local corrections as partial demonstrations and extrapolate them to infer the human’s intended full trajectory for objective learning. \citet{bajcsy2017learning, losey2022physical} interpret human corrections as intentional actions that reveal information about the task objective parameters, allowing the robot to update its model online. \citet{bobu2020quantifying} argue that human corrections may fall outside the robot’s hypothesis space and thus be misleading if incorporated directly. They introduce situational confidence, weighting each correction by how well the robot’s current hypothesis can explain the human input. 

Beyond demonstrations, other studies have modeled corrections as preferences, assuming that the corrected trajectory is more aligned with the human’s desired objective than the original. For example, \citet{jain2015learning, jain2013learning, wilde2020learning} consider corrections as iterative, incremental trajectory refinements that gradually shift the robot’s behavior toward the desired trajectory through comparison-based updates. Similarly, \citet{10.1145/3623384} treat corrections as evidence that the corrected trajectory segment is locally more desirable than nearby perturbations.

However, these works primarily focus on learning from the correction once it has occurred, rather than why or when the human decided to intervene. \citet{korkmaz2025mile} introduced probabilistic models to predict both when an intervention occurs and how the human corrects, integrating these signals into policy learning and achieving high policy update performance. While their approach leverages intervention signals to improve robot behavior, it does not evaluate whether the model itself accurately reflects the underlying human decision-making process. In particular, their framework assumes Boltzmann-rational human actions~\citep{laidlaw2022boltzmann}, where human corrections are modeled probabilistically as being exponentially more likely for actions that yield higher expected rewards. 

While the work above captures behavioral variability, it frames corrections as outcomes of optimization rather than as expressions of human intent shaped by task goals and interaction context. In practice, however, corrections can be indicative of humans’ intent to adjust the robot’s behavior toward a desired outcome: \citet{jin2022learning} infer instantaneous intent from directional feedback, while \citet{schrum2022mind} learn individualized mappings that translate human correction styles into intended objectives. In parallel, Bayesian frameworks for intent recognition in collaborative settings \citep{jain2019probabilistic, iregui2021reconfigurable} demonstrate how probabilistic inference can recover latent human goals from behavioral signals. Together, these insights motivate our approach: if corrections reflect a human’s internal evaluation of the robot’s performance, then their timing and spatial characteristics should provide implicit cues for goal inference.

\subsection{Human Feedback Timing}

In social psychology, \citet{nosek1999response} demonstrates that the timing of a person’s response reveals underlying mental representations and cognitive processes, providing a window into how people internally evaluate and interpret stimuli. Similarly, research on conversational feedback shows that the timing of verbal backchannels reflects a listener’s internal predictive model of dialogue, indicating moments of understanding or alignment with the speaker’s intent \citep{paierl2025distribution}. These findings suggest that the timing of human feedback is not arbitrary but a behavioral manifestation of internal evaluation processes. Extending this idea to human–robot interaction, the timing of human feedback can convey implicit judgments about the robot’s performance. For example, 
our prior work \cite{wang2025effects} examined how human correction feedback is affected by robot behavioral features — such as motion legibility and competency. We found that people tend to intervene earlier and sometimes provide unnecessary corrections when the robot appears more competent. \citet{spencer2022expert} show that the moment of human intervention signals when the robot’s behavior becomes undesirable, effectively partitioning the action space into acceptable and unacceptable regions. Thus, feedback timing serves as a threshold on human tolerance for error and offers an implicit learning signal for the robot.  

While prior work indicates that intervention timing carries valuable information for learning, it does not examine which factors of robot motion drive humans’ intervention decisions or how the timing of those decisions can be leveraged to infer task constraints. We aim to fill this gap by connecting humans' decisions to intervene with early goal inference and precise task constraint learning.

\section{Problem Definition}

Our goal is to model how the timing of a human's interventions can serve as a learning signal for the robot (toward inferring the task goal).
Rather than aim to reproduce human behavior (i.e., behavior modeling), our goal is to infer the underlying task objectives that are indicated by a human's interventions.
We consider a scenario where the robot executes a pre-planned trajectory $\xi = \{x_t\}_{t=1}^T$, where $x_t$ is the robot gripper position at each time step $t$. During this execution, the human may choose to intervene in the robot's motion at any time $t_c \in [1, T]$ to provide a correction. In doing so, they guide the robot through a trajectory $c = \{x_t\}_{t=t_c}^{T_c}$ ($T_c$ being the time when the correction ends), from which we extract timing and spatial information. The timing information consists of $t_c$ (the timestep at which the correction begins). The spatial information consists of $c_p = x_{t_c} \in \mathbb{R}^3$ (position where people grasp the robot gripper), $c_p' = \Delta_t c_p \in \mathbb{R}^3$ (the initial correction direction: velocity), and $c_l= x_{T_c} \in \mathbb{R}^3$ (where people release the gripper at the end of the correction). After the correction, the robot re-plans its trajectory based on its understanding of the task.

To address \textbf{RQ1}, we aim to identify which aspects of robot motion influences the human's decision to intervene.
If certain motion features maximize a model's accuracy to predict correction timing, it suggests that those features are closely tied to the human intervention decision. Our goal is to learn the conditional distribution $\mathbb{P}(t_c \mid \xi, g)$ that models when corrections occur based on the robot original trajectory $\xi$ and task goal $g \in \mathbb{R}^3$ (a gripper position), from which we can quantitatively evaluate which trajectory-based and task-related features explain what triggers intervention decision.

To address whether correction timing information improves direct task goal inference when using correction onset information (\textbf{RQ2}) and enhances precise task goal inference (\textbf{RQ3}), we consider the task goal $g$ drawn from a task-dependent distribution $\mathbb{P}^*(g)$, which represents the likelihood of goal locations that result in successful task completion. This distribution is unknown to the robot and constitutes what it aims to learn. Our objective is to evaluate whether incorporating correction timing $t_c$ can improve the inferred goal distribution $\mathbb{P}(g \mid \xi, t_c, s_c)$ — estimated from the robot’s trajectory $\xi$, timing information $t_c$, and spatial correction cues $s_c$ — to better approximate the true task constraint $\mathbb{P}^*(g)$, compared to inference using only spatial information $\mathbb{P}(g \mid \xi, s_c)$. When $s_c = (c_p, c_p')$, representing the grasp position and initial correction direction, the resulting goal inference evaluates how well the model can infer the goal distribution directly from the onset of correction, addressing \textbf{RQ2}. When $s_c = c_l$, the release position, the model infers the precise goal distribution based on the end of the correction, corresponding to \textbf{RQ3}.

\section{Approach}

We propose a two-stage approach to model and utilize human correction timing for improved robot learning. In the first stage, we train a transformer-based timing prediction model to (i) take task- and motion-related trajectory features as input and (ii) output the probability of the human correcting the robot at each time step. This enables us to analyze which aspects of robot motion prompt human interventions. In the second stage, we train a goal inference model to integrate both the spatial ($s_c$) and temporal ($t_c$) characteristics of the correction to infer the goal distribution $\mathbb{P}(g \mid \xi, t_c, s_c)$. By comparing it against a spatial-only baseline $\mathbb{P}(g \mid \xi, s_c)$, we evaluate whether correction timing offers a useful signal for robot learning; i.e., whether it improves inference of human-intended correction goals and task constraints.

\subsection{Predicting WHEN People Give Corrections}
\subsubsection{Feature Extraction.}
\label{sec:features}

Prior studies have incorporated both velocity-related and task-related motion features into learning frameworks \citep{bajcsy2017learning, jain2020anticipatory}, which are closely tied to what humans perceive and respond to during collaboration. Prior work \cite{dragan2013generating} suggests that a robot's motion may be used to communicate its intent during physical collaboration, and that this communication is improved by optimizing for \emph{legible} motion~\citep{dragan2013legibility}. Characteristics such as human-likeness \citep{maurice2017velocity}, efficiency \citep{jain2020anticipatory}, and safety \citep{ak2023learning, saunders2017trial} influence how easily humans can interpret and adapt to robot motion. The degree to which a robot’s behavior aligns with what humans consider optimal strongly affects their likelihood of intervening \citep{tian2023matters, korkmaz2025mile}. Taken together, these works suggest a robot's motion influences the human's decision to intervene. 

\paragraph{Human Expectation-related Features.} Since humans have a preference for natural robot trajectories \cite{mimnaugh2021defining}, we compute an optimal reference trajectory from each time step to the goal using a PID controller. This trajectory represents a smooth, dynamically feasible motion that a human might perceive as a natural movement toward the goal. From this trajectory, we obtain the \textit{optimal next velocity} $v_t^{\text{opt}}$ and \textit{optimal next position} $x_t^{\text{opt}}$. Additional details about the PID controller and the construction of these reference trajectories are provided in Appendix
\ref{sec: opt pid}.
We then define two alignment-based features: (F1) cosine similarity between the robot’s current velocity $v_t$ and the optimal velocity $v_t^{\text{opt}}$, and (F2) Euclidean distance between the current position $x_t$ and the optimal next position $x_t^{\text{opt}}$.

\begin{table}[t]
\centering
\small 
\setlength{\tabcolsep}{5pt} 
\caption{Time-series features for modeling robot behavior and human corrections, grounded in prior work linking motion characteristics to human perception and intervention.}
\label{table:features}
\begin{tabular}{    p{0.49\columnwidth}
     >{\centering\arraybackslash}p{0.42\columnwidth}}
\toprule
\textbf{Feature} & \textbf{Equation} \\
\midrule
Expectation Alignment (Vel.) \citep{gopinath2020active,gopinath2022information} & 
$\frac{v_t \cdot v^{\text{opt}}_{t}}{\|v_t\| \|v^{\text{opt}}_{t}\|}$ \\
Expectation Alignment (Pos.) \citep{korkmaz2025mile,tian2023matters} & 
$\| x_t - x^{\text{opt}}_t \|$ \\
Directness Alignment (Vel.) \citep{gopinath2020active,gopinath2022information} & 
$\frac{v_t \cdot v^{g}_{t}}{\|v_t\| \|v^{g}_{t}\|}$ \\
Velocity Consistency \citep{xie2019deep,yao2022imitation} & 
$\frac{v_t \cdot v_{t-1}}{\|v_t\| \|v_{t-1}\|}$ \\
Legibility \citep{dragan2013legibility} & 
$\frac{\int \mathbb{P}(g | \xi_{S \rightarrow x_t}) \gamma_t \, dt}{\int \gamma_t \, dt}$ \\
Task Progress (Dist. to Goal) \citep{gopinath2020active,gopinath2022information} & 
$\| x_t - x_{g} \|$ \\
Optimality \citep{bajcsy2017learning, bobu2018learning, bobu2020quantifying, losey2022physical} & 
$\frac{\exp\left(-\mathcal{L}(\xi_{s \rightarrow x_t})-\mathcal{L}^{\text{opt}}(\xi_{x_t \rightarrow g})\right)}{\exp\left(-\mathcal{L}(\xi_{s \rightarrow x_{t-1}})-\mathcal{L}^{\text{opt}}(\xi_{x_{t-1} \rightarrow g})\right)}$ \\
\bottomrule
\end{tabular}
\end{table}

In addition to this locally optimal reference, we also define a direct-to-goal expectation, where the robot is expected to move straight toward the goal without curvature. This yields: (F3) the cosine similarity between the current velocity $v_t$ and a direct velocity vector $v_t^g$ pointing from $x_t$ to the goal $g$.

\paragraph{Dynamics-related Features.} Robot dynamics and velocity patterns impact how human adapt to and interact with the robot ~\citep{xie2019deep, yao2022imitation, maurice2017velocity}. We define (F4) motion consistency, measured by the cosine similarity between \(v_t\) and the previous velocity \(v_{t-1}\).

\paragraph{Task-Performance-Related Features.} To quantify how interpretable the robot’s motion is to a human observer, we included (F5) a legibility score based on~\citet{dragan2013legibility}, shown in Table \ref{table:features} (see
\ref{sec: legi} 
for details). 
Additionally, we capture (F6) task progress as the Euclidean distance from the robot’s current position \(x_t\) to the goal \(g\), following prior work~\citep{gopinath2020active,gopinath2022information}.
And finally, (F7) Boltzmann rational optimality. Prior work~\citep{bajcsy2017learning,bobu2018learning,bobu2020quantifying,losey2022physical} models human intervention likelihood as a function of trajectory optimality, commonly using a Boltzmann rationality model~\citep{ziebart2008maximum,baker2007goal}, where corrections become exponentially more likely as behavior becomes sub-optimal. In our task, we focus on efficiency~\citep{dragan2013legibility} and define optimality as the inverse total path length (executed length to time $t$ plus the optimal remaining path to the goal). 
We represent this as a single time-varying feature: the ratio of optimality between consecutive timesteps (Table~\ref{table:features}).

\subsubsection{Timing Prediction Model.}

\paragraph{Inputs.} To analyze how features evolve over time and how they relate to human intervention, we convert each \emph{original} robot trajectory into a temporal sequence. Each robot pre-planned trajectory $\xi = \{x_t\}_{t=1}^T$ is discretized into $T$ time steps and
converted into a featurized representation $\Phi(\xi, g) = 
\big[\, \phi_t^k(\xi,g) \,\big]_{t=1:T,\,k=1:7} \in \mathbb{R}^{T \times 7}$
with respect to some goal $g$.

\paragraph{Model Architecture.} Given the temporal nature of the data, we adopted a transformer model~\citep{vaswani2017attention} to preserve its sequential structure, which is essential for modeling correction decisions that are inherently non-Markovian~\citep{hochreiter1997long}. For a given trajectory \(\xi\), 
the model takes $\Phi(\xi,g)$ as input, 
applies masking to handle variable-length trajectories, and uses positional encoding to preserve temporal order. The encoder consists of two transformer layers, each comprising a multi-head self-attention mechanism ($8$ heads, embedding dimension $32$) followed by a feed-forward network ($64$ hidden units) with residual connections, dropout, and layer normalization. 

\paragraph{Outputs \& Loss Function.} The final output layer applies a sigmoid activation to produce a corresponding sequence of cumulative distribution function (CDF) probabilities $\mathbb{P}_{\text{CDF}}(t) = \mathbb{P}(t_c \leq t \mid \xi, g) $ representing the likelihood of a correction occurring at or before that time step given the robot trajectory $\xi$ and goal of the task $g$. These predictions are compared against ground truth labels \(\hat{l}_t^i\), where \(\hat{l}_t^i = 0\) if no correction has occurred or the time step precedes the correction, and \(\hat{l}_t^i = 1\) for all time steps following the correction. The model is trained using a binary cross-entropy loss with an exponentially decayed learning rate schedule, and validation loss is monitored to select the best-performing checkpoint.

Given the transformer-predicted CDFs for each trajectory, we can derive the probability density function (PDF) of a correction occurring at each time step $t$ as:
\begin{equation}
\label{eq:correction_timing}
\begin{aligned}
\mathbb{P}(t \mid \xi, g) 
&= \mathbb{P}_{\text{CDF}}(t) - \mathbb{P}_{\text{CDF}}(t-1).
\end{aligned}
\end{equation}
The value of the PDF at the actual observed correction time $t_c$ is taken at $t = t_c$, averaged within a 1.2-second window. Any negative probabilities resulting from numerical artifacts are set to zero.

\subsection{Enhancing Goal Inference}

In this section, we introduce three models for inferring the goal distribution $\mathbb{P}(g \mid \cdot)$ that represents the robot’s estimate of where the goal is. First, we define a \textbf{WHERE model} that relies solely on the spatial information of the correction to infer the goal distribution, serving as a baseline. Next, we present a \textbf{WHEN model} that relies on the timing of human intervention to perform goal inference, isolating the contribution of temporal information. Finally, we present a \textbf{COMBINED model} that integrates both spatial and timing information to infer the goal distribution. By comparing these models, we can evaluate whether the timing of human intervention provides additional informative signals beyond spatial cues, thus testing our hypotheses in \textbf{RQ2} and \textbf{RQ3}.

\subsubsection{Inferring Goal using Timing Information}

Using Eq.~\ref{eq:correction_timing}, the posterior distribution over goals given the correction timing (defined as \textbf{WHEN model} goal distribution) is:
\begin{equation}
\label{eq:pgoal_timing}
\mathbb{P}(g \mid t_c, \xi)  
=\frac{\mathbb{P}(t_c \mid g, \xi)\,\mathbb{P}(g \mid \xi)}
{\sum_{\hat{g} \in \mathcal{G}} \mathbb{P}(t_c \mid \hat{g}, \xi)\,\mathbb{P}(\hat{g} \mid \xi)}
\propto \mathbb{P}(t_c \mid g, \xi),
\end{equation}
assuming our prior over candidate goals $\mathbb{P}(g \mid \xi) = \mathbb{P}(g)$ is uniform; i.e., the robot's pre-planned trajectories are generated independently of the sampled goal hypotheses.

\subsubsection{RQ2: Goal Inference from \emph{Start} of Correction}

We now aim to infer the intended goal location using the spatial information available at the onset of the human correction; specifically, the position where the participant first grasps the gripper $c_p$ and the velocity they apply at that moment $c_p'$. 
Because people do not always release the gripper precisely at the goal after completing the correction, we decompose the process into two stages. By first predicting where participants are likely to release the gripper and then inferring the goal from these predicted endpoints, the model captures the intermediate intent expressed through the correction motion, resulting in a more interpretable and realistic goal inference process.

In the first stage, we infer where people intend to move the robot. 
We implement a feedforward Multilayer Perceptron (MLP) to model $c_l = \text{MLP}(c_p, c_p')$; i.e., using the initial interaction (grasp position $c_p$ and velocity $c_p'$) to predict where the human releases the gripper $c_l$. The network consists of three fully connected layers with hidden dimension 64 and ReLU activations.

In the second stage, we use the predicted release position to infer the goal distribution. We fit a Gaussian Mixture Model (GMM) to model 
the distribution of release positions relative to the ground-truth goal position. This results in a model for  
$\mathbb{P}_{\text{GMM}}(c_l \mid g)$.

Combining both the MLP and GMM, we can approximate the \textbf{WHERE model} goal distribution as follows (derivation in \ref{sec: grasp math}):
\begin{equation}
\mathbb{P}(g \mid c_p, c_p') \approx \mathbb{P}_{\text{GMM}}(\text{MLP}(c_p, c_p') \mid g),
\end{equation}

\subsubsection{Combining WHEN and WHERE}

To leverage both spatial and temporal information, 
we compute the posterior distribution over candidate goals conditioned on the robot trajectory $\xi$, correction timing $t_c$, and the spatial information available at the onset of the correction, including the grasp position $c_p$ and the initial correction velocity $c_p'$ (the \textbf{COMBINED model}):

\begin{align}
\mathbb{P}(g \mid t_c, c_p, c_p', \xi) 
&= 
\frac{
\mathbb{P}(t_c \mid g, \xi) \, \mathbb{P}_{\text{GMM}}(\text{MLP}(c_p, c_p') \mid g)
}{
\sum\limits_{\hat{g} \in \mathcal{G}} 
\mathbb{P}(t_c \mid \hat{g}, \xi) \, 
\mathbb{P}_{\text{GMM}}(\text{MLP}(c_p, c_p') \mid \hat{g})
}.
\label{eq:pg_combined}
\end{align}
See \ref{sec: combined} for the full derivation.
Following Eq.~\ref{eq:pg_combined}, we also experiment with weighing $\mathbb{P}_w(t_c \mid g, \xi)$ and $\mathbb{P}(c \mid g)$ differently, according to weight $\alpha\in[0,1]$. 
The \textbf{weighted COMBINED model} becomes:
\begin{equation}
\label{eq:weighted_combined}    
\mathbb{P}_w(g \mid t_c, c_p, c_p', \xi) = 
\frac{
\mathbb{P}(t_c \mid g, \xi)^{\alpha} \cdot \mathbb{P}_{\text{GMM}}(\text{MLP}(c_p, c_p') \mid g)^{1-\alpha} 
}
{
\sum\limits_{\hat{g} \in \mathcal{G}} \mathbb{P}(t_c \mid \hat{g}, \xi)^{\alpha} \cdot \mathbb{P}_{\text{GMM}}(\text{MLP}(c_p, c_p') \mid \hat{g})^{1-\alpha}
}.
\end{equation}

\subsubsection{RQ3: Goal Inference from \emph{End} of Correction}

Additionally, we investigate whether timing information can enhance the precision of goal distribution inference. Instead of using the grasp position $c_p$ and its corresponding velocity $c_p'$, we directly utilize previously fitted GMM $\mathbb{P}_{\text{GMM}}(c_l \mid g)$. The resulting posterior for the \textbf{WHERE model} is $\mathbb{P}(g \mid c_l) \propto \mathbb{P}_{\text{GMM}}(c_l \mid g)(c_l \mid g)$.
The \textbf{weighted COMBINED model} becomes:
\begin{equation}
\label{eq:weighted_combined_release}    
\mathbb{P}_w(g \mid t_c, c_l, \xi) = 
\frac{
\mathbb{P}(t_c \mid g, \xi)^{\alpha} \cdot \mathbb{P}_{\text{GMM}}(c_l \mid g)^{1-\alpha}
}
{
\sum\limits_{\hat{g} \in \mathcal{G}} 
\mathbb{P}(t_c \mid \hat{g}, \xi)^{\alpha} \cdot \mathbb{P}_{\text{GMM}}(c_l \mid \hat{g})^{1-\alpha}
}.
\end{equation}

\section{Evaluation}

In our prior work~\cite{wang2025effects}, we conducted a user study to collect correction data with $N=120$ participants recruited from a university community, resulting in a total of 7,435 interaction episodes and 3,585 correction trajectories. We now use this data to train and test our models in an offline manner\footnote{Data-processing code available at 
\href{https://github.com/iqr-lab/correction-timing}{\nolinkurl{https://github.com/iqr-lab/correction-timing}}.}.

\subsection{Data Collection}

\paragraph{Study Setup.} Each participant was tasked with supervising a 7-DoF Kinova Gen3 robotic arm equipped with a Robotiq 2F-85 gripper, mounted on a horizontal linear actuator. During each participant's 1-hour session, the robot performed a series of 64 pick-and-place operations, where the goal was to insert various shapes into matching color-coded target holes (goals). For the block placement task, we consider four distinct shapes — circle, square, triangle, and rectangle — each paired with four colored holes positioned at different locations on the board. Participants were free to intervene in the robot's motion at any time in order to provide a correction. 

To incentivize high-quality data, participants were told that the robot was learning from their feedback in real-time, and that they would receive a bonus compensation based on the number of successful robot trials. In reality, the robot followed pre-determined waypoints based on the participant’s study condition\footnote{For full information about the study, refer to our prior work \cite{wang2025effects}.} (rather than learning in real-time), and participants received the base + maximum bonus compensation (as if the robot had succeeded at every trial) to ensure that they were fairly compensated regardless of their study condition. We obtained approval for this study through our Institutional Review Board and followed ethics protocol for debriefing participants on these hidden elements.

\paragraph{Pre-Planned Robot Motion.}
The robot approached each sub-task by executing motion pre-planned with RRT*~\cite{sucan2012the-open-motion-planning-library}, which were smoothed before being executed through a velocity-based PID controller. To pre-plan trajectories, we assigned target goal poses that were intentionally varied in their correctness (e.g., they may be slightly off target or correspond to the wrong color target) and optimized trajectories according to different legibility levels defined by our study conditions\footnotemark[3]. We did this to reflect how a robot may be deployed with an imperfect task policy, and how these failures and inefficiencies prompt humans' subjective decisions to intervene and correct the robot.

\paragraph{Physical Interactions.}

Upon applying physical force to interrupt and modify the robot's motion, an admittance control scheme~\cite{kim2019model, jenamani2024feel} allowed the robot to respond compliantly. We used the recursive Newton-Euler algorithm for inverse dynamics and gravity compensation~\cite{carpentier-rss18}. Corrections terminated when participants ceased applying force; the robot then replanned from the corrected state to the closest viable goal while preserving the end-effector’s final orientation. Sensor readings from joint encoders, torque sensors, and control inputs were collected at 10Hz. Cartesian states are computed through forward dynamics~\citep{coumans2016pybullet}.

\begin{figure}
    \centering
    \includegraphics[width=\linewidth]{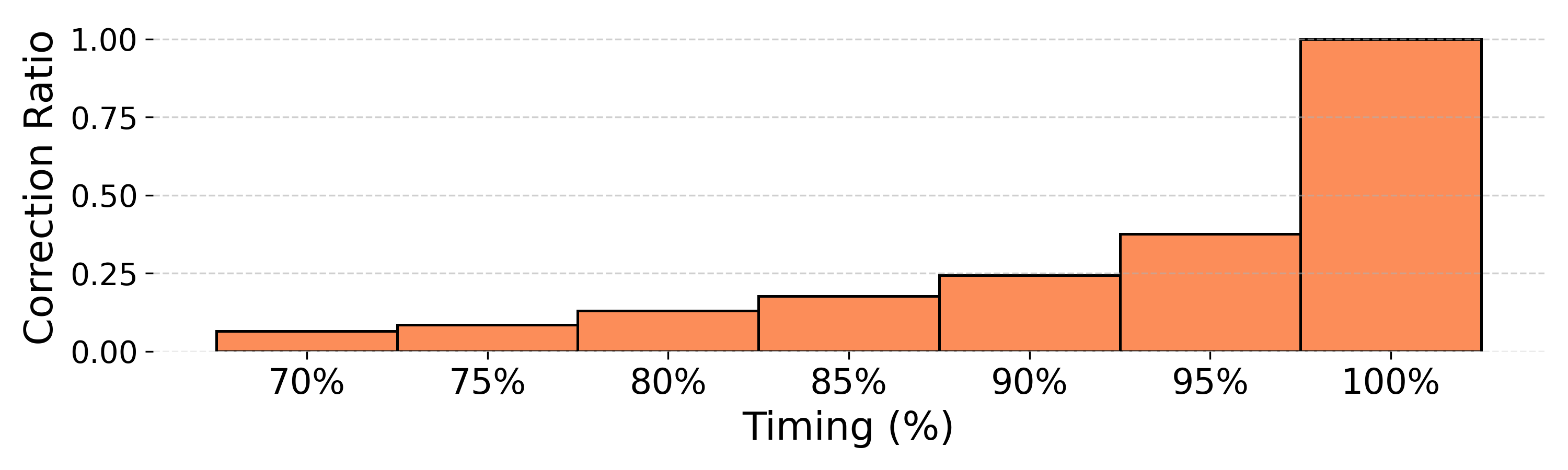}
    \caption{Cumulative first-correction timing across trajectory completion. “Correction Ratio” denotes the proportion of first-correction events occurring up to each completion percentage. 
}
    \label{fig:correction_histo}
\end{figure}

Although participants could issue multiple interventions, we restrict our analysis to the \emph{first correction} event in each trial to keep the problem tractable. We find that these first corrections occur throughout different phases of the robot’s trajectory (Fig.~\ref{fig:correction_histo}), meaning that our study covers early and late corrections.

\subsection{Model Training}

Because our model operates over full trajectories, we construct a unified representation for both corrected and non-corrected trials. For non-corrected trials, we use the uninterrupted executed trajectory. For corrected trials, we combine the pre-correction executed segment with the intended (but unexecuted) remainder of the pre-planned trajectory.

We first create sets corresponding to correction timing percentages ($\leq$70\%, $\leq$80\%, $\leq$90\% and $\leq$100\%), measured as the proportion of the trajectory executed prior to the first correction. We did not analyze earlier corrections due to their sparsity, comprising only 6.5\% of the data. For each percentage set, we split the trajectories involving human corrections into a training set (60\%), a validation set (10\%), and a test set (30\%). For correction timing prediction, we trained the transformer model using features extracted from the training and validation sets, and included an equal number of uncorrected trajectories during training to enable the model to predict both \emph{if} and \emph{when} corrections occur. For goal distribution inference, we trained the MLP and fitted the GMM using the same training and validation sets with only the trajectories involving corrections. Each unique shape–color pairing defines a distinct task (goal) constraint that the model is trained to learn. 

\subsection{Evaluation Metrics}

For \textbf{RQ1}, we evaluate the accuracy of correction timing prediction and analyze the contribution of each trajectory-based feature through ablation studies. For \textbf{RQ2} and \textbf{RQ3}, we assess the WHEN model’s ability to enhance the precision of goal inference.

\subsubsection{Transformer Accuracy}
We evaluate the model’s correction timing prediction accuracy to assess whether the trajectory-based features effectively capture the motion factors that drive human intervention. Strong prediction performance indicates that these features are informative and relevant for answering \textbf{RQ1} — understanding what aspects of robot motion lead to human decisions to intervene. We compare performance of our multi-feature model against a single-feature Boltzmann baseline, where the only input to the transformer is the Boltzmann optimality feature $\Phi_t = [\phi_t^{7}]$, representing the commonly used model for correction behavior in prior work. All evaluations are conducted over 200 random training, validation, and test splits of the dataset to ensure that the observed performance is robust and statistically reliable. All sets include equal numbers of correction-inducing and uncorrected trajectories.

(1) \textbf{F1 Score}: We report the mean F1 score on the test set to evaluate timestep-level correction prediction. 
Predicted probabilities $p_t$ are thresholded at $0.5$ to obtain binary labels $l_t = H(p_t - 0.5)$, which are compared against ground-truth labels, where $H(\cdot)$ denotes the unit step function.

(2) \textbf{Correction MAE}: To evaluate correction timing accuracy, we define the predicted correction time as the first timestep where $l_t$ transitions from $0$ to $1$ and remains $1$ thereafter. 
For each trajectory, we compute the absolute difference between the predicted time $t_c$ and ground-truth time $\hat{t}_c$. We report the mean absolute error (MAE) over the test set: $\text{MAE} = \frac{1}{N} \sum_{i=1}^{N} \left| \hat{t}_c^{(i)} - t_c^{(i)} \right|$.

(3) \textbf{Predicted Correction Ratio}: We evaluate intervention detection accuracy by computing the ratio of predicted corrected trajectories to the actual number of corrected trajectories: $\frac{N_{\text{pred-corr}}}{N_{\text{true-corr}}}$ in each test set.

\subsubsection{Feature Importance}

We further analyze the contribution of each feature to address \textbf{RQ1}. We conduct a feature ablation study, excluding one feature at a time during training and evaluating the resulting drop in F1 score. This analysis reveals which aspects of the robot’s motion most influence human correction timing decisions at different trajectory stages. The evaluation is conducted over 200 random training, validation, and test splits of the dataset.

\subsubsection{Goal Inference Accuracy}

We investigate whether timing information improves goal inference using information from the start of corrections (\textbf{RQ2}) and end of corrections (\textbf{RQ3}).
For all models, we uniformly sample candidate goal positions $\hat{g}$ on the $(x, y)$ plane at $z = 0$, assuming that the $z$-coordinate has minimal influence on participants' perception of the goal. The sampled region spans $40\,\text{cm} \times 60\,\text{cm}$ around the board area where the actual goals are located, with a sampling resolution of $\Delta x = \Delta y = 1\,\text{cm}$.

We compare the mean Kullback–Leibler Divergence (KLD)~\citep{kullback1951information} between the inferred and ground truth goal distributions for the \textbf{WHEN}, \textbf{WHERE}, and weighted \textbf{COMBINED} models across all trajectories in the test set to evaluate their goal inference accuracy. For the weighted COMBINED model, we set $\alpha = 0.8$ to balance the contributions of timing and spatial information (details about how we picked $\alpha$ in \ref{sec: alpha}). The evaluation is conducted over 50 random training, validation, and test splits of the dataset.

\paragraph{Ground Truth.} 
To estimate the ground truth goal distribution for each shape, we employ a sim-to-real approach~\cite{da2025survey}. We first conduct block-dropping simulations~\citep{coumans2016pybullet} across a range of potential target positions, recording if the block lands in the target. Using eigen-entropy \citep{huang2023eigen}, we select a set of 100 poses that were maximally informative about the simulators, execute the poses in the real world, and recorded the outcomes. Comparing simulated and real results identifies the most accurate simulator for each shape.

Using the selected simulators, we performed $10^5$ additional block drops at randomly sampled poses for each shape. Successful placements were aggregated to construct the ground-truth goal distribution, modeled as a GMM centered at the true target positions and evaluated at $z = 0.08$, since small $z$ variations do not affect success. 
For each colored target, we applied an offset to align the GMM center with the corresponding absolute target position.

\section{Results}

\begin{figure}
    \centering
    \includegraphics[width=\linewidth]{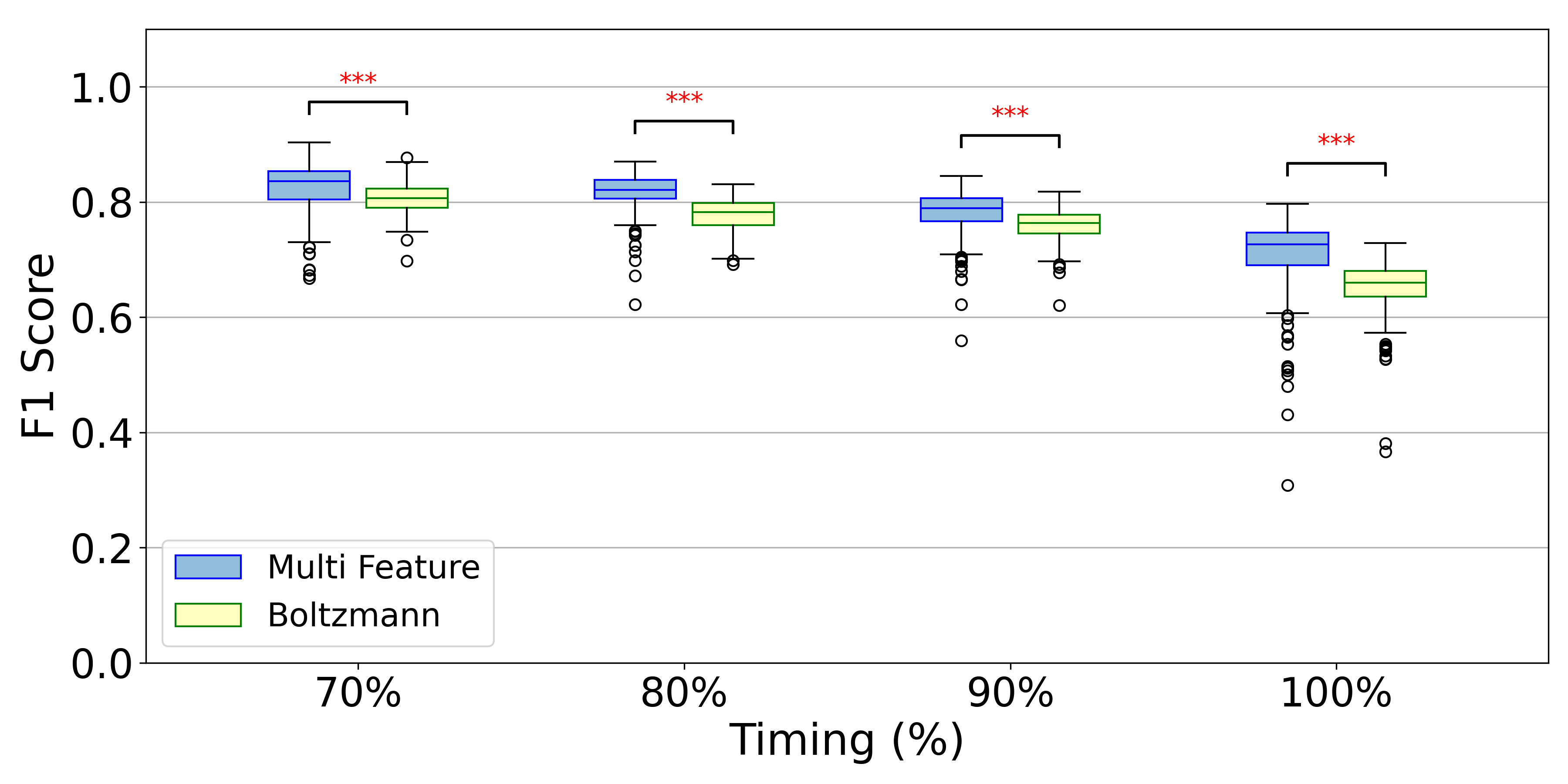}
    \caption{F1 scores for the multi-feature model and Boltzmann baseline across correction timing percentages (portion of trajectory completed before the first correction). Statistical significance in all figures is indicated as $p \leq 0.05^*$, $p \leq 0.01^{**}$, $p \leq 0.001^{***}$.}
    \label{fig:f1_all}
\end{figure}

\begin{figure}
    \centering
    \includegraphics[width=\linewidth]{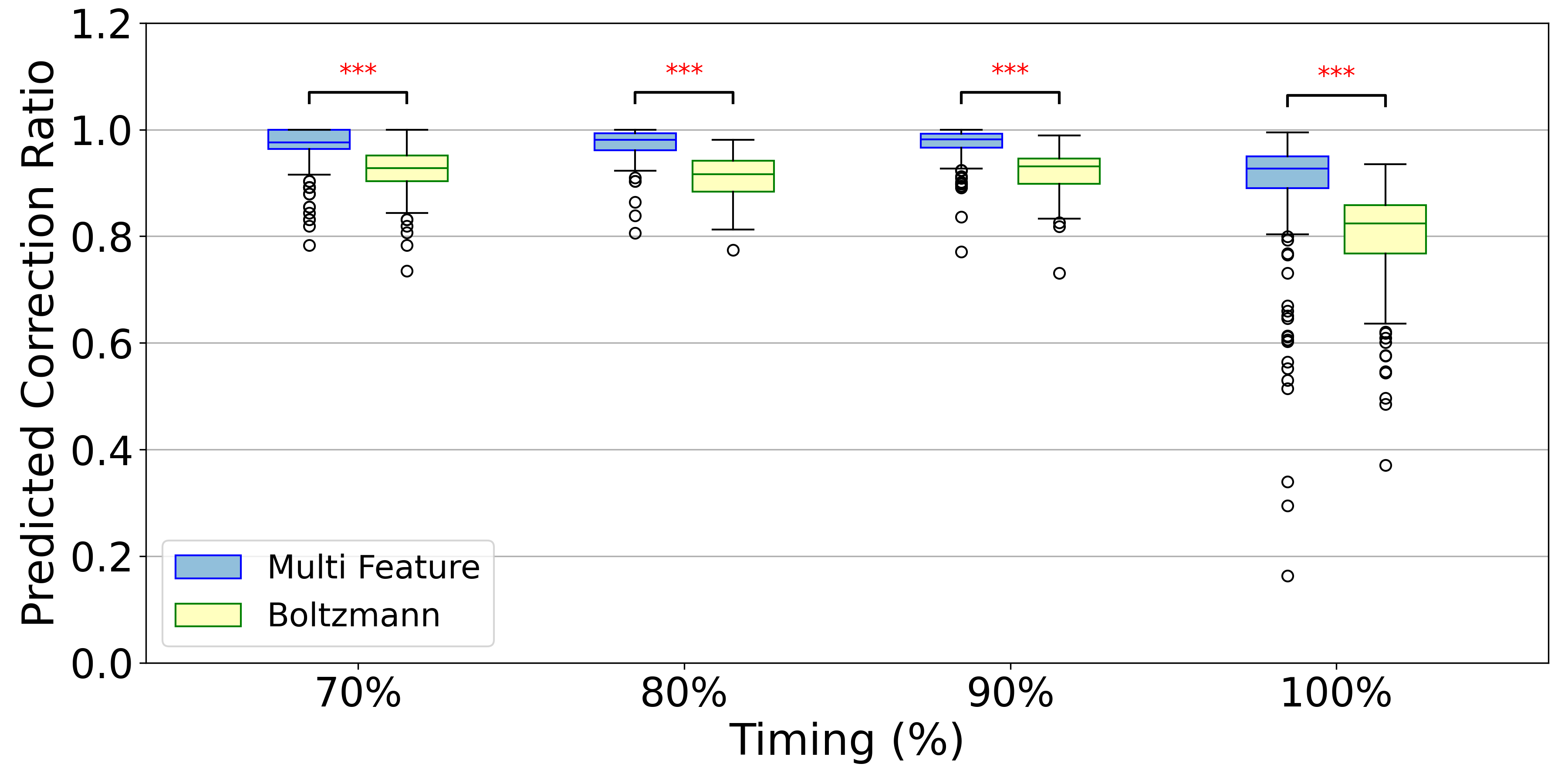}
    \caption{Predicted-to-actual correction ratio for the multi-feature model and Boltzmann baseline across correction timing percentages.}
    \label{fig:prediction_ratio_all}
\end{figure}

\begin{figure}
    \centering
    \includegraphics[width=\linewidth]{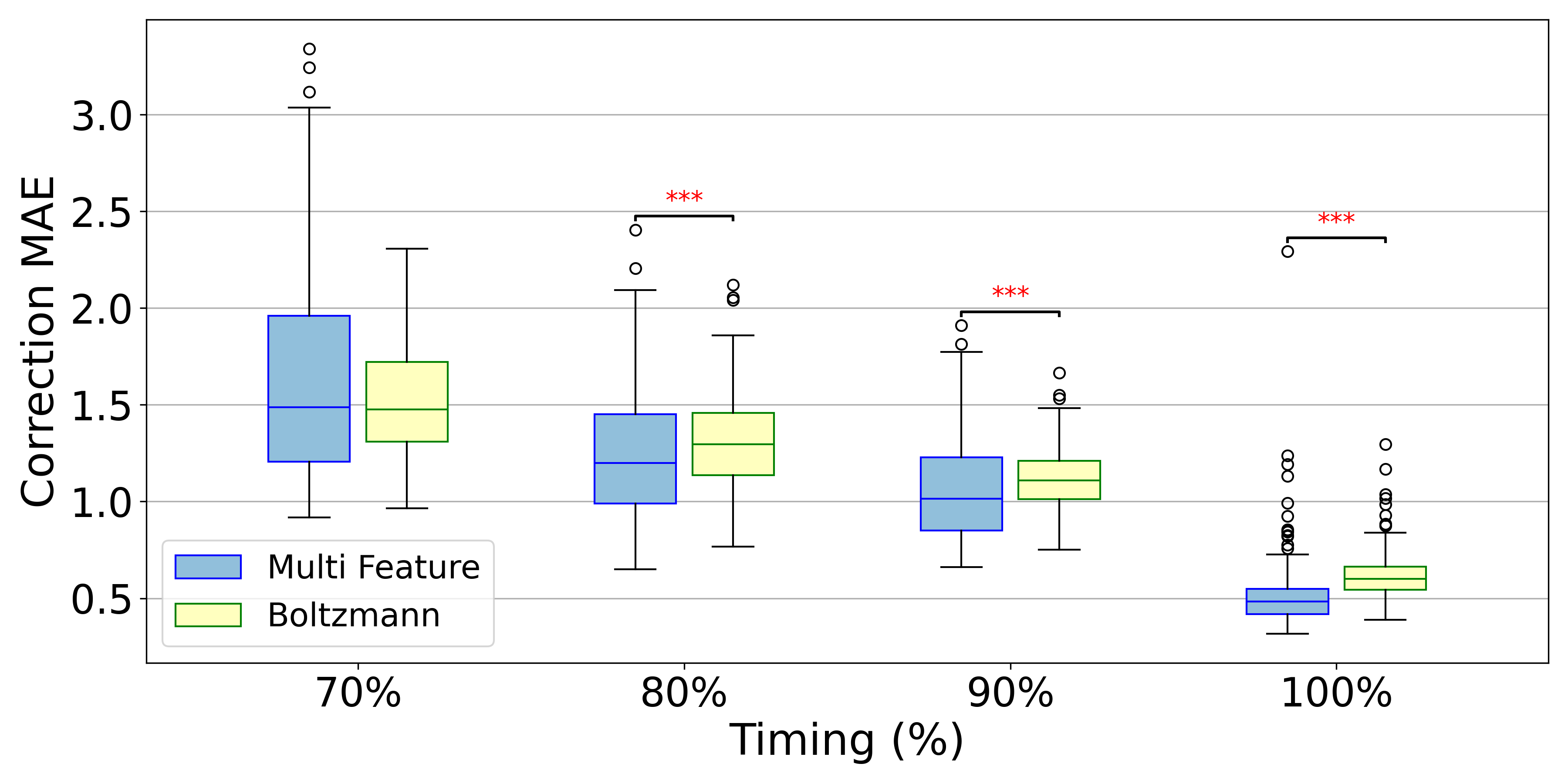}
    \caption{Mean absolute error (seconds) between predicted and true correction timing for the multi-feature model and Boltzmann baseline across correction timing percentages. Lower MAE indicates better accuracy.}
    \label{fig:correction_mae_all}
\end{figure}

\begin{figure}
    \centering
    \includegraphics[width=\linewidth]{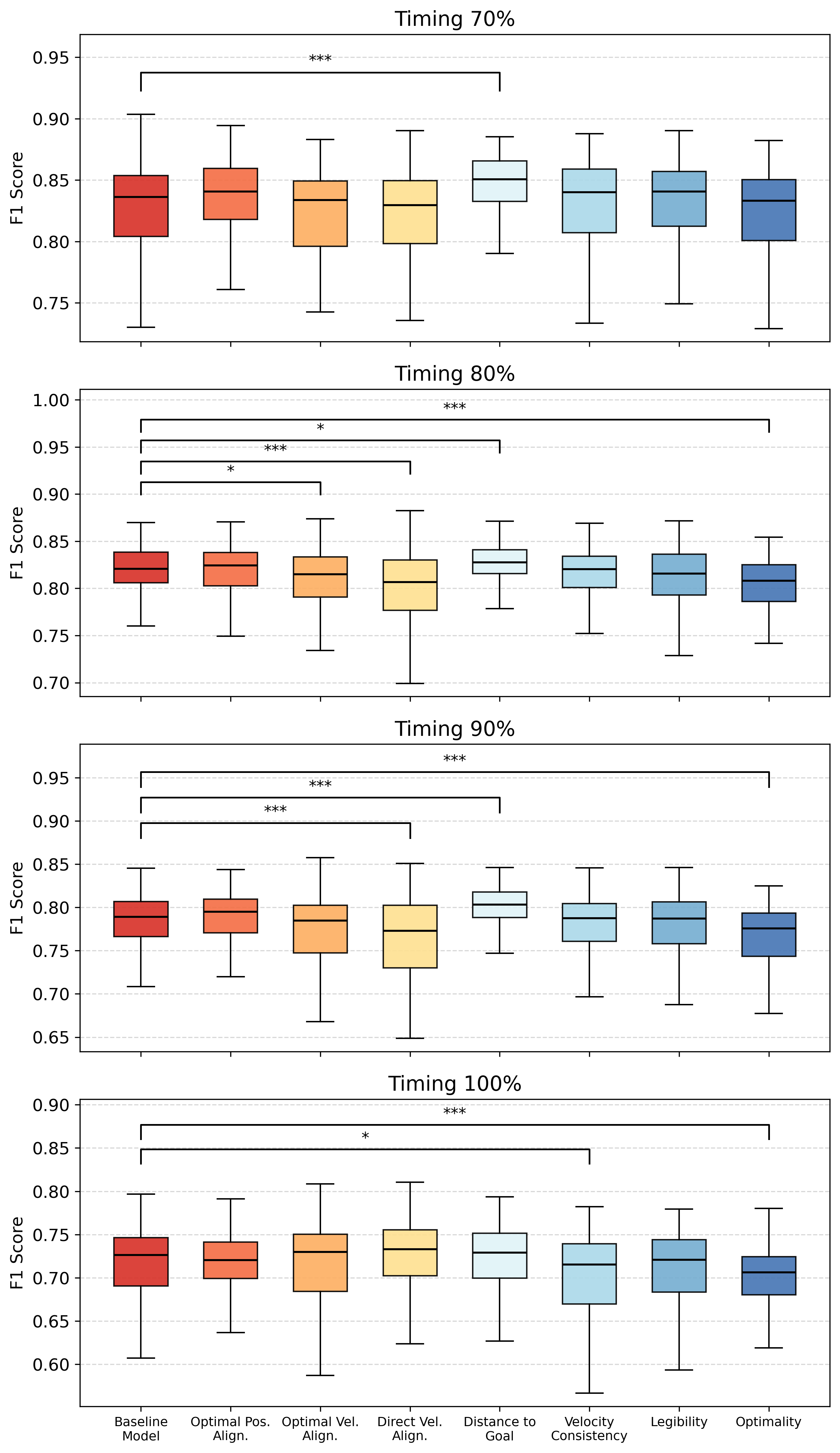}
    \caption{F1 scores for the baseline and single-feature ablations across correction timing percentages. Significance denotes deviation from the baseline, indicating feature impact. A drop in F1 suggests positive contribution of the removed feature.}
    \label{fig:feature_importance}
\end{figure}

\begin{figure*}
     \centering
     \begin{subfigure}[b]{0.49\textwidth}
         \centering
         \includegraphics[width=\textwidth]{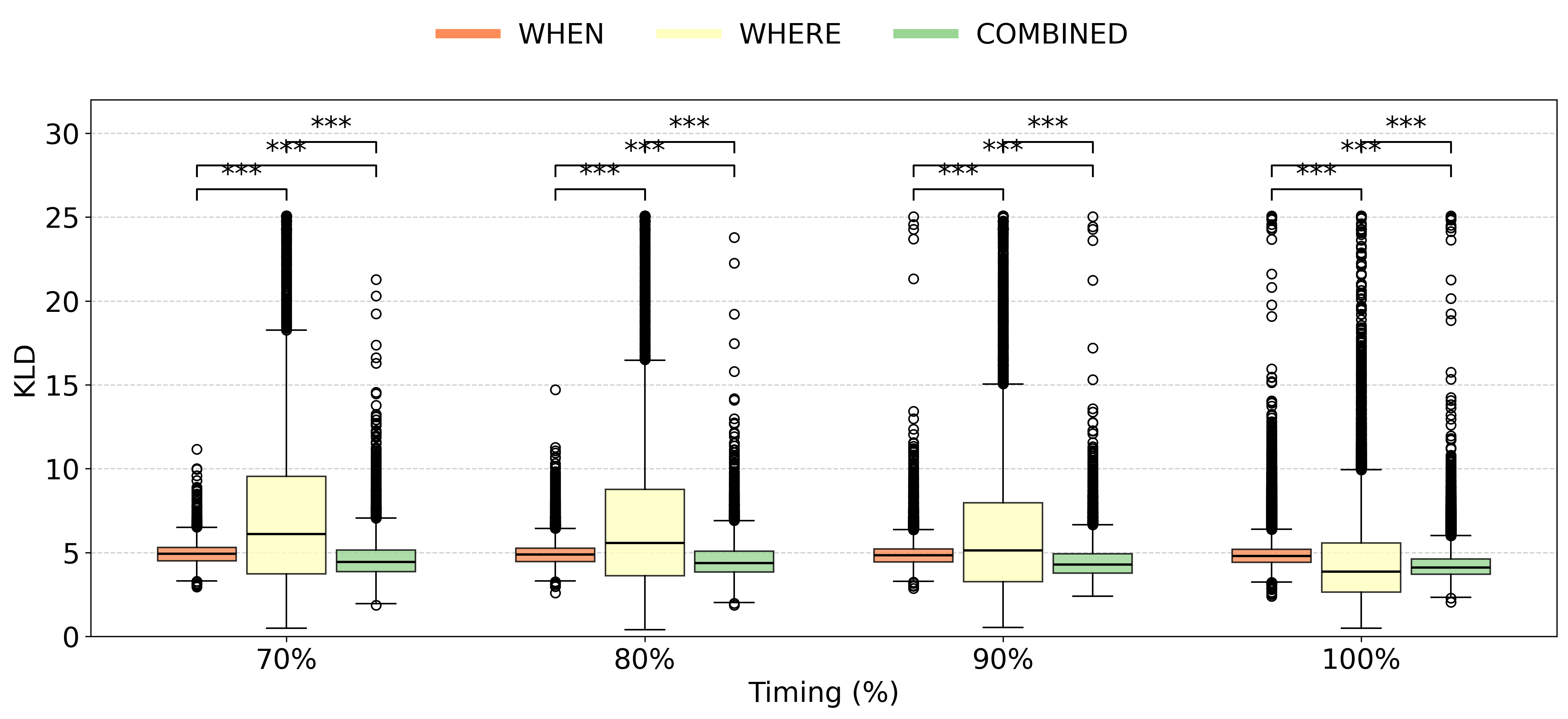}
         \caption{Results of using \emph{start}-of-correction data to infer goal distribution.}
         \label{fig:KLD_grasp}
     \end{subfigure}
     \hfill
     \begin{subfigure}[b]{0.49\textwidth}
         \centering
         \includegraphics[width=\textwidth]{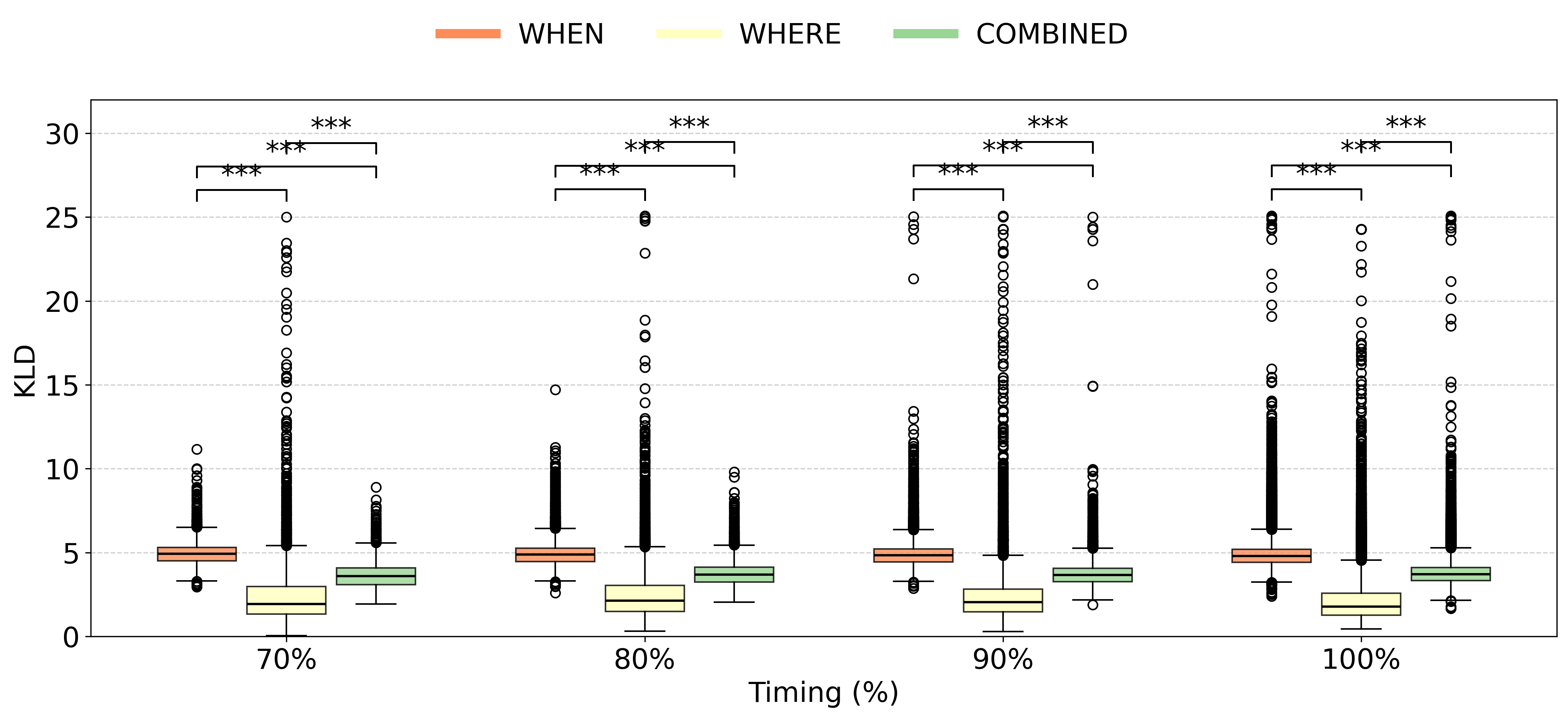}
         \caption{Results of using \emph{end}-of-correction data to infer goal distribution.}
         \label{fig:KLD_release}
     \end{subfigure}
        \caption{KLD between ground-truth and inferred goal distributions after each correction in the test set, aggregated over 50 random splits. Results are shown for each goal inference model (WHEN, WHERE, COMBINED) across correction timing percentages, aggregated over targets and shapes. Small circles denote outliers beyond $1.5\times$ the interquartile range.}
        \label{fig:KLD_combined}
\end{figure*}

(1) \textbf{Correction Timing Accuracy}: 
Across 200 runs, the multi-feature model consistently outperformed the Boltzmann baseline in F1 score, particularly when later corrections were included (Fig. \ref{fig:f1_all}). It also achieved better correction timing accuracy at 80\%, 90\% and 100\% of the trajectory (Fig. \ref{fig:correction_mae_all}), and was significantly more accurate in predicting the number of actual corrections across all correction timing percentages (Fig. \ref{fig:prediction_ratio_all}). At the same time, both models showed declining F1 scores and predicted/actual correction ratios as later corrections were considered, while the correction timing becomes more accurate.

(2) \textbf{Feature Importance}: 
For feature importance in Fig. \ref{fig:feature_importance},
we observed that different trajectory features affected the model’s F1 score at different stages of the trajectory, mostly resulting in a decrease when the feature was removed. Some feature removals led to minimal or insignificant changes. Boltzmann optimality removal led to performance drops at 80\%, 90\%, and 100\%. Optimal velocity alignment removal decreased performance at 80\%, direct velocity alignment at 80\% and 90\%, and velocity consistency when corrections occurring at the end of the trajectory (100\%) were included. However, distance to goal removal led to a noticeable increase in F1 score at 70\%, 80\%, and 90\% of the trajectory.

(3) \textbf{Goal Inference from \emph{Start} of Correction}:
As shown in Fig.~\ref{fig:KLD_grasp}, when using the grasping location $c_p$ and initial velocity $c_p'$ to infer the goal distribution, the COMBINED model achieves significantly lower KLD than either the WHEN or WHERE models alone for each actual target before the latest corrections are included (70\%, 80\%, 90\%). However, when later corrections are included (100\%), the COMBINED model performs worse than the WHERE model.

(4) \textbf{Goal Inference from \emph{End} of Correction}: 
As shown in Fig.~\ref{fig:KLD_release}, when using the leaving location $c_l$ to infer the goal distribution, the COMBINED model achieves lower KLD than the WHEN model but does not outperform the WHERE model across all correction timing percentages.

\section{Discussion}

\textbf{Can we predict intervention timing using robot motion features?}
We demonstrate that the model successfully predicts both \emph{whether} and \emph{when} a correction will occur, achieving strong F1 scores (Fig.~\ref{fig:f1_all}), high correction capture ratios (Fig.~\ref{fig:prediction_ratio_all}), and low mean absolute timing errors (Fig.~\ref{fig:correction_mae_all}). These results indicate that human intervention decisions are non-Markovian — they depend not only on the current state but also on the preceding trajectory history. Moreover, our multi-feature model consistently outperforms the Boltzmann baseline, particularly when later corrections are included, suggesting that human intervention decisions are influenced by multiple aspects of robot motion rather than a single optimality-based factor.

\textbf{Which features of robot motion influence intervention decisions?}
The feature ablation results (Fig.~\ref{fig:feature_importance}) show that multiple motion features jointly influence correction prediction, with no single feature causing a drastic performance drop. Removing optimality consistently reduces performance at 80\%, 90\%, and 100\%, indicating its stable relevance for later-stage corrections. Direct velocity alignment has the strongest impact at 80\% and 90\%, suggesting it shapes mid-trajectory intervention decisions. Velocity consistency affects performance only at 100\%, aligning with the robot’s natural deceleration near task completion, while optimal velocity alignment matters mainly at 80\%. Interestingly, removing distance-to-goal improves performance, suggesting that humans attend less to absolute goal distance and more to motion cues — such as alignment and efficiency — that signal whether the robot behaves as expected.

\textbf{Can timing information improve inference of human intended goals and task objectives?}
We find that when using information available at the onset of a correction, the COMBINED model achieves higher goal inference accuracy than both the WHEN and WHERE models alone for earlier corrections (70\%, 80\%, and 90\%) (Fig.~\ref{fig:KLD_grasp}). This indicates that timing information enhances early goal inference from correction onset before the correction completes.

In contrast, when goal inference is based on the release position of the gripper, the COMBINED model does not outperform the WHERE model (Fig.~\ref{fig:KLD_release}). In our dataset, participants typically released the gripper very close to the intended goal, making the release position itself an almost complete indicator of the goal distribution. Consequently, timing information adds little value for refining task objectives in this setting.

\textbf{In what settings it useful to incorporate timing for learning from corrections?} Correction timing enhances prediction of participants’ intended correction goals when combined with initial contact information, enabling earlier inference of intent without observing the full correction. Building on shared autonomy frameworks like \citet{javdani2018shared}, such early intent prediction could help robots use feedback more efficiently and provide proactive assistance as soon as an intervention begins.

Although timing does not improve task constraint inference from correction endpoints, this likely reflects our task’s simplicity; in more complex settings with less direct correction–goal mappings, timing may remain a valuable complementary signal.

\textbf{When are human corrections most informative?}
The timing prediction model performs best for mid-trajectory corrections, showing higher F1 scores (Fig.~\ref{fig:f1_all}) and correction detection rates (Fig.~\ref{fig:prediction_ratio_all}). In our dataset, many interventions happen very late (Fig.~\ref{fig:correction_histo}), with participants often waiting until the end of the task to intervene. These late-stage corrections are harder to predict due to the limited number of positive samples (less time steps before correction occurs) and provide less informative feedback as much of the task has already been executed. The smaller mean timing error at late stages (Fig.~\ref{fig:correction_mae_all}) likely results from the greater number of late corrections.

Consistent with this, early goal inference performs best for earlier corrections, whereas the benefit of timing information diminishes for late corrections (100\%) (Fig.~\ref{fig:KLD_grasp}). Since participants often grasp the robot near the goal, the WHERE model already predicts the intended endpoint accurately, leaving little room for timing cues to help. Additionally, with minimal trajectory remaining, late corrections provide limited opportunity for timing-based inference.

\textbf{Limitations and Future Work.} We found no improvement in goal inference from post-correction positions, likely because our task was simple and direct — the end position already revealed the goal. Future work will explore how to learn from corrections in more complex tasks involving multiple constraints.
While our model predicts whether and when corrections occur, it has yet to be integrated into real-time planning. Embedding this predictive ability into online control could help robots anticipate human goals and adapt dynamically.
Finally, our current feature set, though effective, may not capture all factors driving human corrections; richer, task-specific features could enhance future models.

\section{Conclusion}

Our results show that correction timing provides meaningful insight into human intervention behavior and offers valuable information for early goal inference. We successfully model both whether and when humans issue corrections, showing that these decisions depend on the trajectory history rather than the current state alone. Feature ablation analysis indicates that multiple aspects of robot motion jointly influence intervention timing. Based on this, integrating timing with spatial cues in the COMBINED model improves goal inference accuracy, enabling earlier prediction of human intended correction goal and supporting more proactive robot assistance. However, timing adds little benefit when goal inference is based on the gripper’s final release position, as this already provides a near-complete signal of the task goal. Overall, correction timing is most valuable for early intent inference when spatial information is incomplete, highlighting its role as a complementary signal for learning from human feedback.



\begin{acks}
The authors would like to thank Haimanot Belachew, Ulas Berk Karli, Ziyao Shangguan, Anushka Potdar, Jirachaya “Fern” Limprayoon, Rebecca Ramnauth, Nicholas C. Georgiou, Debasmita Ghose, Liam Hoffmeister-Ricke, Sasha Lew and Paul Comeau for their feedback, contributions and support.
\end{acks}



\bibliographystyle{ACM-Reference-Format} 
\bibliography{sample}


\clearpage

\appendix

\section{Appendix}

\subsection{Feature Details}

\subsubsection{Optimal Trajectories}
\label{sec: opt pid}

We generate a reference trajectory using a PID controller following the original pre-planned trajectory to obtain the optimal velocities \(v_t^{\text{opt}}\) and optimal positions \(x_t^{\text{opt}}\) used to compute the alignment-related features.
At each time step \(t\) along an original trajectory, the controller produced a “natural” trajectory from the robot’s current state—specifically its position \(x_t\) and velocity \(v_t\)—toward the goal \(G^*\). We consider this reference trajectory to be “optimal” in the sense that it reflects what a human might intuitively expect the robot to do from that point onward, based on the smoothness and predictability of PID-generated motion~\citep{maurice2017velocity}. We then compute the position and velocity by interpolating the dynamics 0.25 seconds into the future along this reference trajectory to represent \(x_t^{\text{opt}}\) and \(v_t^{\text{opt}}\), respectively.

\subsubsection{Legibility}
\label{sec: legi}

To represent legibility, we adopt the formulation from~\citet{dragan2013legibility} and compute a score $l_t$ at each time step \(t\) as:

\begin{equation}
l_t =  \frac{\int \mathbb{P}(G^* \mid \xi_{S \rightarrow x_t}) \, \gamma_t \, dt}{\int \gamma_t \, dt},
\label{eq:appendix_legibility}
\end{equation}

where \(\gamma_t = T - t\) is a discount factor that gives more weight to earlier parts of the trajectory and \(T\) is the total trajectory duration. The term \(\mathbb{P}(G^* \mid \xi_{S \rightarrow x_t})\) is the probability that the robot is going to goal \(G^*\), given the observed trajectory prefix \(\xi_{S \rightarrow x_t}\), and is computed as:

\begin{equation}
\mathbb{P}(G^* \mid \xi_{S \rightarrow x_t}) \propto 
\frac{
\exp\left( - \mathcal{C}(\xi_{S \rightarrow x_t}) - \mathcal{C}(\xi^\text{opt}_{x_t \rightarrow G^*}) \right)
}{
\exp\left( - \mathcal{C}(\xi^\text{opt}_{S \rightarrow G^*}) \right)
}
\mathbb{P}(G^*).
\label{eq:goal_inference}
\end{equation}

This probability measures how efficient it is to reach the goal \(G^*\) via the current observed trajectory snippet \(\xi_{S \rightarrow x_t}\), followed by an optimal completion \(\xi^\text{opt}_{x_t \rightarrow G^*}\), relative to the cost of the optimal trajectory \(\xi^\text{opt}_{S \rightarrow G^*}\). In our implementation, we define the cost function \(\mathcal{C}(\cdot)\) as the total trajectory length. Higher legibility indicates that the robot's motion strongly suggests its intention to reach the goal \(G^*\) early on.

\subsection{Derivation of the Goal Inference Formulation}
\subsubsection{Inferring from Where People Grasp the Gripper}
\label{sec: grasp math}

First we derive the posterior over goals given the release position is
\begin{equation}
\label{eq:pgoal_release}
\small
\mathbb{P}(g \mid c_l) \propto \mathbb{P}(c_l \mid g) \cdot \mathbb{P}(g) \propto \mathbb{P}(c_l \mid g),
\end{equation}
assuming a uniform prior over goals.

During testing, for each trajectory, given the grasp position $c_p$ and velocity $c_p'$, the MLP predicts the release position:
\begin{equation}
\hat{c}_l = f_{\text{MLP}}(c_p, c_p').
\end{equation}

We assume $c_l$ and $\hat{c_l}$ follow the same distribution. Through the MLP and using the Baye's Rule from Eq. \ref{eq:pgoal_release}, the posterior over goals conditioned on the grasp position and velocity becomes (defined as \textbf{WHERE model} goal distribution):
\begin{equation}
\mathbb{P}(g \mid c_p, c_p') = \mathbb{P}(g \mid \hat{c_l}) \propto \mathbb{P}(\hat{c_l} \mid g) \approx \mathbb{P}_{\text{GMM}}(\text{MLP}(c_p, c_p') \mid g),
\end{equation}
where $\mathbb{P}(\hat{c_l} \mid g)$ is calculated using the GMM fitted for $\mathbb{P}(c_l \mid g)$.

\subsubsection{Combining When and Where}
\label{sec: combined}

To derive the posterior over goals using both spatial and timing information:
\begin{equation}
\label{eq:bayes_general}
\mathbb{P}(g \mid t_c, c_p, c_p', \xi) = 
\frac{\mathbb{P}(t_c, c_p, c_p', \xi \mid g) \mathbb{P}(g)}
{\mathbb{P}(t_c, c_p, c_p', \xi)}.
\end{equation}

We expand the joint probability in the numerator using the chain rule:
\begin{equation}
\mathbb{P}(t_c, c_p, c_p', \xi \mid g)
= \mathbb{P}(t_c, c_p, c_p' \mid g, \xi)\,\mathbb{P}(\xi \mid g).
\end{equation}

In our setup, the robot’s pre-planned trajectories are generated independently of the sampled goal hypotheses. Therefore, we assume that the trajectory $\xi$ and the goal $g$ are independent a priori, such that  $\mathbb{P}(\xi \mid g) = \mathbb{P}(\xi)$.
Substituting this into Eq.~\ref{eq:bayes_general} gives:
\begin{equation}\small
\mathbb{P}(g \mid t_c, c_p, c_p', \xi) = 
\frac{\mathbb{P}(t_c, c_p, c_p' \mid g, \xi)\,\mathbb{P}(g)}
{\mathbb{P}(t_c, c_p, c_p' \mid \xi)}.
\label{eq:independence_applied}
\end{equation}

We assume conditional independence between the correction timing $t_c$ and the correction initiation parameters $(c_p, c_p')$ given the goal $g$ and the trajectory $\xi$, as they capture distinct aspects of the human intervention process: \emph{when} the human decides to intervene, whereas $(c_p, c_p')$ describe \emph{how} the correction is initiated spatially and dynamically.:
\begin{equation}
\mathbb{P}(t_c, c_p, c_p' \mid g, \xi) = \mathbb{P}(t_c \mid g, \xi)\,\mathbb{P}(c_p, c_p' \mid g, \xi).
\end{equation}
This assumption is supported empirically in our correlation analysis (Appendix~\ref{sec: correlation tc cp}).

Substituting back:

\begin{equation}
\small
\mathbb{P}(g \mid t_c, c_p, c_p', \xi)  \\
= \frac{
\mathbb{P}(t_c \mid g, \xi) \mathbb{P}(c_p, c_p' \mid g, \xi) \mathbb{P}(g)
}
{\mathbb{P} (t_c, c_p, c_p' \mid \xi)}.
\end{equation}

The denominator is a normalization constant computed as the sum over all possible goals $\mathcal{G}$.

\begin{equation}
\small
\mathbb{P}(t_c, c_p, c_p' \mid \xi) = 
\sum_{\hat{g} \in \mathcal{G}} \mathbb{P}(t_c \mid \hat{g}, \xi) \mathbb{P}(c_p, c_p' \mid \hat{g}, \xi) \mathbb{P}(\hat{g}).
\end{equation}

Thus, we have

\begin{equation}
\small
\mathbb{P}(g \mid t_c, c_p, c_p', \xi) = 
\frac{
\mathbb{P}(t_c \mid g, \xi)  \mathbb{P}(c_p, c_p', \mid g, \xi) \mathbb{P}(g)
}
{
\sum\limits_{\hat{g} \in \mathcal{G}} \mathbb{P}(t_c \mid \hat{g}, \xi)  \mathbb{P}(c_p, c_p' \mid \hat{g}, \xi) \mathbb{P}(\hat{g}).
}
\end{equation}

To calculate $\mathbb{P}(c_p, c_p' \mid \hat{g}, \xi)$, we assume that $c_p, c_p'$ is independent of $\xi$ since the correction initiation should reflect human's intent based on the task goal rather than the shape of the robot's pre-planned trajectory, so it becomes $\mathbb{P}(c_p, c_p' \mid \hat{g})$. This assumption is partially supported by our correlation analysis in Appendix \ref{sec: correlation cp xi}, where the grasp position $c_p$ exhibits negligible correlation with trajectory similarity across different $\xi$’s. $c_p'$ is dependent on where the correction begins, so the weak dependence of $c_p$ suggests that the influence of $\xi$ on correction initiation can be reasonably ignored for our modeling purposes.

Eventually, the equation becomes:

\begin{equation}
\small
\mathbb{P}(g \mid t_c, c_p, c_p', \xi) = 
\frac{
\mathbb{P}(t_c \mid g, \xi) \mathbb{P}(c_p, c_p', \mid g) \mathbb{P}(g)
}
{
\sum\limits_{\hat{g}\in \mathcal{G}} \mathbb{P}(t_c \mid \hat{g}, \xi) \mathbb{P}(c_p, c_p' \mid \hat{g}) \mathbb{P}(\hat{g})
}.
\end{equation}

Potential goals are uniformly sampled, thus $P(\hat{g})$ is uniform, so the posterior (\textbf{COMBINED model} goal distribution posterior) becomes:

\begin{equation}
\label{eq:combined}    
\mathbb{P}(g \mid t_c, c_p, c_p', \xi) = 
\frac{
\mathbb{P}(t_c \mid g, \xi) \mathbb{P}(c_p, c_p' \mid g) 
}
{
\sum\limits_{\hat{g} \in \mathcal{G}} \mathbb{P}(t_c \mid \hat{g}, \xi) \mathbb{P}(c_p,  c_p' \mid \hat{g})
},
\end{equation}

where $\mathbb{P}(t_c|g, \xi)$ is derived from Eq.\ref{eq:correction_timing}, and through the MLP:
\begin{equation}
\label{eq:p_grasp_goal}
\mathbb{P}(c_p, c_p' \mid g) \propto \mathbb{P}(\hat{c}_l \mid g) = \mathbb{P}_{\text{GMM}}(\text{MLP}(c_p, c_p') \mid g).
\end{equation}
Thus,
\begin{align}
\mathbb{P}(g \mid t_c, c_p, c_p', \xi) 
&= 
\frac{
\mathbb{P}(t_c \mid g, \xi) \, \mathbb{P}_{\text{GMM}}(\text{MLP}(c_p, c_p') \mid g)
}{
\sum\limits_{\hat{g} \in \mathcal{G}} 
\mathbb{P}(t_c \mid \hat{g}, \xi) \, 
\mathbb{P}_{\text{GMM}}(\text{MLP}(c_p, c_p') \mid \hat{g})
}.
\label{eq:pg_combined}
\end{align}

\subsection{Correlations}

\subsubsection{Correlation Between Correction Timing $t_c$ and Spatial Information ($c_p, c_p'$)}
\label{sec: correlation tc cp}

To verify the assumption that correction timing is largely independent of spatial displacement, we computed the Pearson correlation between the normalized correction timing (in seconds) and the grasp positions relative to the actual goal and the initial correction velocity along each spatial dimension. Across all 3,589 filtered correction samples, we observe the following correlations for the grasp position $c_p$ and the initial correction velocity $c_p'$:

\begin{table}[h]
\centering
\caption{Pearson correlation between normalized correction timing $t_c$ and correction spatial features.}
\resizebox{\columnwidth}{!}{%
\begin{tabular}{lccc}
\toprule
 & Corr. w/ $x$ & Corr. w/ $y$ & Corr. w/ $z$ \\
\midrule
Grasp position $c_p$ & 0.092 & 0.039 & -0.600 \\
Initial correction velocity $c_p'$ & 0.077 & -0.010 & 0.264 \\
\bottomrule
\end{tabular}%
}
\end{table}

The results show weak correlations between $t_c$ and both $x$ and $y$ components of $c_p$ and $c_p'$, suggesting that the timing of intervention is largely independent of where and how participants initiate the correction in the horizontal plane. The moderate correlations along the $z$-axis likely reflect the vertical component of the motion, which is peripheral to goal inference that primarily occurs in the $x$–$y$ plane. Therefore, we approximate $t_c$ and $(c_p, c_p')$ as conditionally independent given $g$ and $\xi$.

\subsubsection{Correlation Between Grasp Position $c_p$ and Trajectory $\xi$ Similarities}
\label{sec: correlation cp xi}

To verify that the spatial similarity between grasp locations is independent from the overall trajectory similarity, we computed both dynamic time warping (DTW)~\cite{senin2008dynamic} distances between full correction trajectories and Euclidean distances between their corresponding correction endpoints. From 3,589 correction trajectories, we randomly sampled 353,500 trajectory pairs and calculated the Pearson and Spearman correlations between the two distance measures.

The Pearson correlation coefficient was $r = 0.02$ ($p < 0.001$) and the Spearman rank correlation was $\rho = 0.04$ ($p < 0.001$), both indicating negligible relationships between trajectory and endpoint similarities. These results confirm that trajectories that are temporally or kinematically similar do not necessarily result in spatially similar correction endpoints. Therefore, it is reasonable to treat the temporal and spatial components of human corrections as independent in our modeling framework.

\subsection{Model Details}

\subsubsection{Transformer}

We optimize binary cross-entropy over valid time steps only, using the mask $M$ as sample weights. We use Adam~\citep{kingma2014adam} with gradient clipping ($\lVert g\rVert_2$ clipnorm $=1.0$) and an exponentially decaying learning rate ($\eta_0{=}10^{-3}$, decay\_steps $=1000$, decay\_rate $=0.9$). Model selection is done via a held-out validation set by monitoring validation loss and saving the best checkpoint. Random seeds are fixed for reproducibility. Input features are rescaled with the same preprocessing on train/validation splits.

\subsubsection{MLP}

The model is trained using mean squared error (MSE) loss and optimized with the Adam~\citep{kingma2014adam} optimizer with an exponentially decayed learning rate. Validation loss is monitored at each epoch, and the best-performing model is saved based on the lowest validation loss. All inputs and outputs are normalized using statistics computed from the training set to ensure stable convergence.

\subsection{Geometric Weight $\alpha$}
\label{sec: alpha}

\begin{figure}
    \centering
    \includegraphics[width=\linewidth]{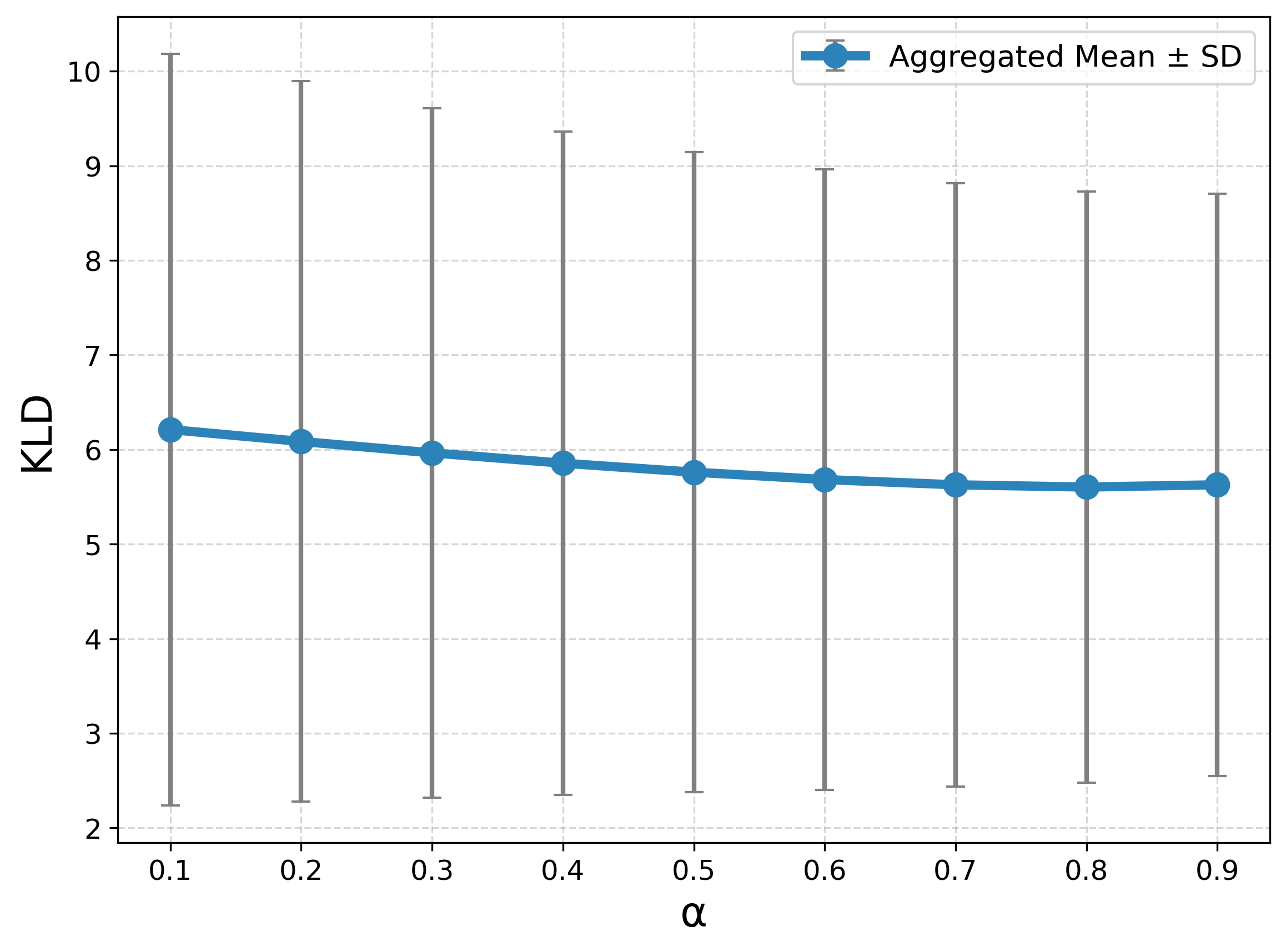}
    \caption{KLD vs. different weight $\alpha$}
    \label{fig:kld_alpha}
\end{figure}

To determine the balance between timing and spatial information in the COMBINED model, we varied 
$\alpha \in [0.1, 0.9]$ in the weighted posterior:
\begin{equation}
\label{eq:weighted_combined}    
\mathbb{P}_w(g \mid t_c, c_p, c_p', \xi) = 
\frac{
\mathbb{P}(t_c \mid g, \xi)^{\alpha} \cdot \mathbb{P}_{\text{GMM}}(\text{MLP}(c_p, c_p') \mid g)^{1-\alpha}
}
{
\sum\limits_{\hat{g} \in \mathcal{G}} \mathbb{P}(t_c \mid \hat{g}, \xi)^{\alpha} \cdot \mathbb{P}_{\text{GMM}}(\text{MLP}(c_p, c_p') \mid \hat{g})^{1-\alpha}
}.
\end{equation}

For testing, we evaluated how the KLD between the inferred and ground-truth goal distributions (using correction onset information $c_p, c_p'$) varies with $\alpha$ at 80\% correction timing percentages, aggregated across all targets and shapes over 5 runs. The results in Fig.~\ref{fig:kld_alpha} show that $\alpha = 0.8$ yields the lowest mean KLD. Therefore, we set $\alpha = 0.8$ as the weighting parameter for all COMBINED models.

\subsection{KLD Plot for Different Targets Using Correction Onset Information}

\begin{figure}[H]
    \centering
    \includegraphics[width=\linewidth]{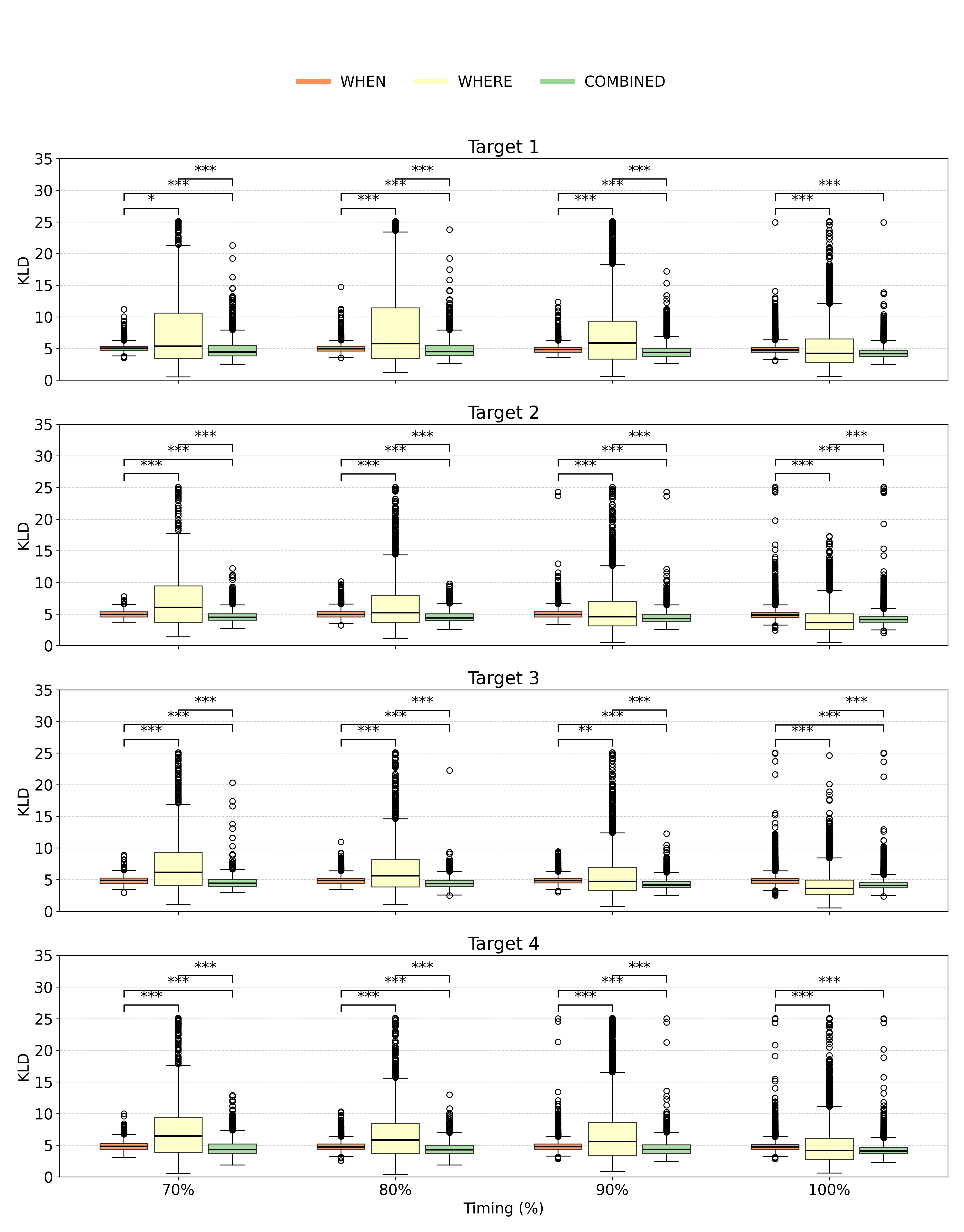}
    \caption{KLD between the ground-truth goal distributions and the inferred goal using the start of the correction information after each correction in the test set, evaluated over 50 random dataset splits. Results are shown for each goal inference model (WHEN, WHERE, and COMBINED) across different correction timing percentages. The results are aggregated across shapes.}
    \label{fig:kld_grasp_by_target}
\end{figure}

We evaluate goal inference performance at each target using the grasp location $c_p$ and initial velocity $c_p'$ as inputs. As shown in Fig.~\ref{fig:kld_grasp_by_target}, the COMBINED model yields significantly lower KLD than either the WHEN or WHERE models for all targets when earlier corrections are considered (70\%, 80\%, 90\%). When late corrections are included (100\%), the COMBINED model performs comparably to WHERE for side targets (targets 1 and 4; no significant difference) but slightly worse for center targets (targets 2 and 3). These results suggest that timing information is most beneficial for early corrections and becomes less informative when corrections occur late and target distinctions are less pronounced.

\end{document}